\documentclass{bmvc2k}
\usepackage{booktabs}
\usepackage{multirow}
\usepackage{caption}
\usepackage{amssymb}

\title{Sparse Competition during Training\\ For the Emergence of Specialized Modules}

\addauthor{Baptiste Rossigneux\textsuperscript{1}}{baptiste.rossigneux@inria.fr}{1}
\addauthor{Karim Haroun\textsuperscript{2}}{karim.haroun@univ-paris8.fr}{2}

\addinstitution{
 \makebox[0pt][r]{\textsuperscript{1}\,}Univ. Rennes\\
 Inria, IRISA\\
 Rennes, France
}
\addinstitution{
 \makebox[0pt][r]{\textsuperscript{2}\,}University of Paris 8\\
 LIASD\\
 Paris, France
}

\runninghead{B. Rossigneux, K. Haroun}{Sparse Competition for Specialized Modules}

\begin{document}

\maketitle

\begin{abstract}
Modularity in deep neural networks has been proposed as a means of improving both interpretability and training by promoting disentangled representations and reducing redundancy. In this work, we study the emergence of modular structure through competition dynamics between groups of neurons during training. We introduce a method that (i) maintains near-baseline accuracy, (ii) induces usage-based modularity by sparsely routing inputs to neuron groups, and (iii) encourages specialization of these modules, such that their activations are correlated with input classes. We evaluate the proposed approach on ImageNet-100 and CIFAR-100 and show that with it, specialized modules emerge without module-level supervision. These modules capture a meaningful high-level structure in the data, with individual modules responding to semantic categories (e.g., dogs or vehicles). We also study the emergence of a hierarchical partition of sub-tasks depending on the number of modules. Our results suggest that competitive dynamics can serve as a simple mechanism for inducing functional modularity in standard architectures.
\end{abstract}

\section{Introduction}
\label{sec:intro}

Modular computation has long been viewed as a promising direction towards more interpretable and reusable neural representations. Early work on Mixture-of-Experts and sparse routing \cite{jacobs1991adaptive, Shazeer2017OutrageouslyLN} proposed architectural solutions to enforce module boundaries. However, architectural separation does not guarantee that modules acquire meaningful functional roles. A neural network can be made of modules while still representing task-relevant information in a redundant or entangled way.

Recent mechanistic interpretability work \cite{elhage2022toy, bricken2023monosemanticity} reveals that networks decompose computation into identifiable sub-circuits. However, these studies reveal the existence of an internal structure without asking whether it is well-organized. Do networks learn reusable functions or redundant and cluttered heuristics that solve the task? The field's focus on benchmark performance obscures this distinction. A neural network can achieve high accuracy through reusable localized computations or opaque distributed heuristics.

Subsequent work suggests that internal representations depend not only on the architecture and the task, but also on the training process. 
For example, open-ended evolutionary systems, such as Picbreeder can produce networks with reusable internal regularities and meaningful responses to perturbations \cite{secretan2008picbreeder,secretan2011picbreeder,huizinga2018emergence}. 
This motivates our question: can standard network training be biased toward more factored internal organization?

A natural approach is to enforce modularity directly. The recent work on clusterability \cite{golechha2025studyingcrossclustermodularityneural} proposes adding a differentiable penalty to the loss that encourages weights to cluster within pre-defined module groups. This achieves structural modularity because the weight matrices become concentrated within the boundaries of the module. However, this comes at a cost. Enforcing clusterability damages the accuracy of the task and fails to induce semantic specialization, caused by the modules not developing meaningful task-related roles. This results in a tradeoff between accepting reduced accuracy for modular weights and abandoning the structural approach altogether. Therefore, can we achieve both modularity and strong task performance?

In this paper, we propose that modularity should emerge from learned routing, not from weight structure. We introduce a routing-based approach that uses module dropout with dual entropy objectives. The first objective concentrates each input on a few modules, while the second encourages different inputs to use different modules. Modules compete for input, which drives specialization without explicit architectural constraints. On ImageNet-100, sparse top-1 routing remains within roughly one percentage point of the vanilla baseline while inducing strong semantic specialization. Classes with the same coarse label are routed to the same modules approximately $3\times$ more often than expected by chance. This specialization also persists hierarchically as increasing module count refines partitions while preserving semantic coherence.

Moreover, we observe that dense training shows specialized modularity under the same competitive incentives. Yet, sparse training forces modules to be individually deployable, whereas dense training creates co-adapted specialists that rely on each other. This distinction separates the emergence of modular structure from the emergence of useful sparse modularity. The representation structure can be induced just by the training dynamics. Furthermore, we extend our approach in several directions. First, we propose learning module widths through a shared capacity budget to create competition over resources that drive specialization. Second, we apply module routing to multiple layers simultaneously and reveal how semantic hierarchy emerges across depths.

\let\thefootnote\relax\footnote{Code is available at: https://github.com/BabaVegato/Minimal-SCM}

Our main contributions break down as follows:

\begin{itemize}
    \item We introduce Sparse Competitive Modules (SCM), a label-free routing objective that induces usage-based modularity through per-sample sharpness and population-level balance.
    \item We show that the resulting modules acquire specialization, with module assignments carrying substantially more class information than without modules or with clustered baselines.
    \item We demonstrate that sparse competitive training is mechanistically necessary for sparse deployment: dense specialization is not enough.
    \item We analyze the semantic organization of learned modules across module counts and layers, showing that learned partitions refine hierarchically while preserving semantic coherence.
\end{itemize}




\section{Related Works}

Most of the recent efforts on modularity tackle either interpretability or efficiency. The interpretability goal consists of decomposing a model into weakly interacting components, making its internal functioning easier to interpret and analyze, while efficiency is achieved through conditional computation. Early work on modular representations in deep neural networks \cite{watanabe2017modular} attempts to produce a simplified description of networks through the use of similar connection patterns. Subsequent work formalizes this approach through the notion of clusterability \cite{filan2021clusterabilityneuralnetworks}, where some parts of the network are thought of as clusters because its parts are densely connected, while they are sparsely connected to the rest. Following this, \cite{golechha2025studyingcrossclustermodularityneural} measures clusterability in a differentiable manner, making it possible to train a neural network on this as an objective, through a loss function that encourages non-interacting clusters. They show that this produces modular components in fully connected layers, but find that these clusters do not reliably become more task-specialized than those in non-modular baselines. Our work addresses this gap. Instead of enforcing modularity as a static property of the weight graph, we study whether sparse competitive routing can induce usage-based modules that are preferentially selected by different visual classes. Moreover, recent work \cite{bena_dynamics_2025} in controlled neural systems shows that structural modularity alone is insufficient for functional specialization and that specialization might depend on resource constraints and architecture. Our work argues that usage-based modularity is sufficient to attain specialization in a neural network. Gradient routing \cite{cloud2024gradient} instead localizes capabilities by supplying data-dependent gradient masks. SCM differs in that module assignments are not specified externally: they emerge from label-free competitive routing.

As models scale to billions or more recently to trillions of parameters, Mixture-of-Experts (MoEs) ~\cite{Shazeer2017OutrageouslyLN,Fedus2021SwitchTS,lepikhin2021gshard} have emerged as a reliable approach to scaling in the architectural design space. MoEs consist of a routing function that activates only a subset of parameters for each input, allowing practitioners to increase model capacity without increasing the computation. In computer vision, V-MoE shows that sparse expert routing can be competitive with dense Vision Transformers while reducing inference cost \cite{Riquelme2021ScalingVW}. Similarly, Soft-MoE shows that it is possible to design fully differentiable MoE architectures to avoid hard token routing~\cite{puigcerver2023sparse}. MoEs also emerged in efficient vision architectures such as CNNs \cite{Zhang_2023_ICCV}, showing that experts can be coupled with tightly vision-prior based architectures. Our goal differs from most MoE work. We investigate the relevance of adding experts as an architectural prior that constrains the network's exploration of the space and its expressivity. We show that softer steering through the loss function can lead to their emergence. This unsupervised emergence of modules is also interesting for interpretability reasons, as it allows us to study the learning dynamics and how they evolve compared to models with enforced modularity through the architecture.

Our routing objective is also closely related to information-maximization clustering objectives such as RIM~\cite{Gomes2010DiscriminativeCB} or Invariant Information Clustering~\cite{ji2019invariant} which balance confident assignments and diversity maintenance across a population. The distinction is in the use of the information-maximization: SCM uses not to learn a clustering head or explicit router, but to induce a specialization inside a neural network layer.

Finally, our work is related to mechanistic interpretability methods that seek to design or identify circuits from deep neural networks as human-understandable units \cite{cunningham2023sparse,bricken2023monosemanticity}. Although we do not introduce new circuit-discovery algorithms, we argue that our contribution to training for emergent modular specialization paves the way for constructing circuits in an unsupervised yet human-understandable manner.

 \begin{figure}[!t]
\begin{center}
\includegraphics[width=\linewidth]{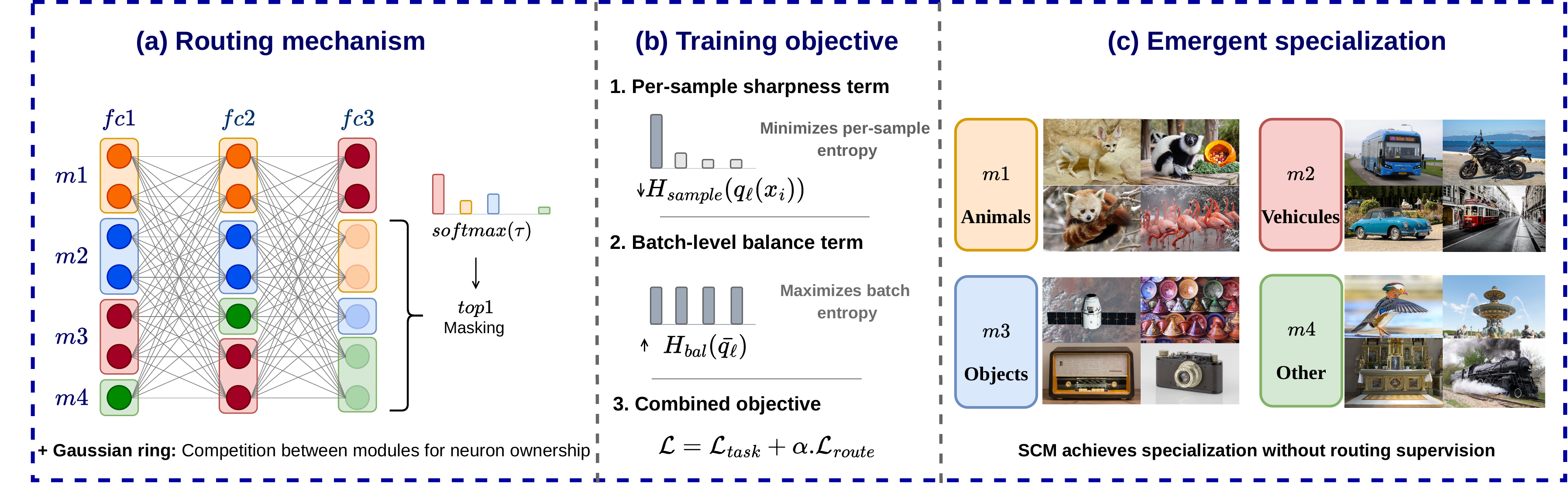}
\end{center}
   \caption{
Overview of Sparse Competitive Modules. 
(a) A routed layer is partitioned into modules. For each input, module activation energies are converted into routing probabilities, and a hard top-1 mask keeps only the selected module active. 
(b) The routing objective combines a per-sample sharpness term, which encourages confident module selection, with a batch-level balance term, which prevents collapse by encouraging all modules to be used. 
(c) Without module-level or semantic supervision, the resulting routing assignments become class-informative, where visually related classes tend to share dominant modules.
}
\label{fig:explanation}
\end{figure}

\section{Methodology}

This section introduces \emph{Sparse Competitive Modules}, for which an overview is presented in Figure~\ref{fig:explanation}, a training procedure that induces usage-based modularity through the loss function of neural networks. Unlike structural modularity objectives, our approach does not directly form weakly interacting clusters. Instead, it steers the single-module selection per input while preventing collapse. Only one module is activated per routed layer, and the objective encourages the network to balance usage across modules. Importantly, the routing decision is unsupervised by class labels, which are used only in the standard classification loss.

\subsection{Modularizing a Layer}
\label{sec:modularizing_layer}

Let \(h_\ell(x)\in\mathbb{R}^{d_\ell}\) be the activation of a routed layer $\ell\in\mathcal{R}$. We divide its coordinates into \(K_\ell\) modules \(\mathcal{G}_{\ell,1},\ldots,\mathcal{G}_{\ell,K_\ell}\). In the simplest case, these are equal-width contiguous blocks of neurons. This partition defines where modules are, but does not assign any semantic role.

For each module $k$, let $m_{\ell,k}\in\{0,1\}^{d_\ell}$ be the
corresponding binary mask. The activation supported by the module $k$ is

\begin{equation}
h_{\ell,k}(x)=m_{\ell,k}\odot h_\ell(x),
\qquad
h_\ell(x)=\sum_{k=1}^{K_\ell} h_{\ell,k}(x).
\end{equation}

We route inputs using the activation energy of each module,
\begin{equation}
    E_{\ell,k}(x) = \left\| m_{\ell,k}\odot h_\ell(x) \right\|_2 + \varepsilon ,
    \label{eq:module_energy}
\end{equation}
where \(\varepsilon\) is a small numerical constant stability. A module is selected on the basis of the magnitude of its response to the input. Unlike structural modularity objectives, which impose constraints directly on the weight matrices, our method leaves the weights unconstrained and induces modularity through input-dependent usage patterns. We later relax the equal-width block assumption by using a Gaussian-ring, described at the end of the next subsection.

\subsection{Sparse Competitive Routing Objective}

The goal of our method is to induce structured module usage without providing any class-related information. For each training sample \((x_i, y_i)\), the label $y_i$ is used only through the standard classification loss. The routing objective does not observe $y_i$, it operates only on the activation patterns produced by the network itself. Hence if visual classes become associated with different modules, this structure emerges from competition between modules driven by input-dependent usage, rather than from any externally specified partition of the data.

Our objective combines two competing pressures. The first encourages each input to make a confident routing decision, localizing its computation to a small subset of modules. The second encourages balanced usage across inputs, ensuring that the network retains the capacity of all modules. This balance is necessary because sparse routing has a natural collapse mode that causes all inputs to be routed to the same module. Given $k$ modules of equal size, this would effectively reduce the usable capacity of the routed layer to roughly $1/k$ of its original capacity. Collapse can be self-reinforcing: a slightly more responsive module receives more routed examples, hence more task gradients, which increase its dominance. Similarly, if one module initially learns a broadly useful classification strategy, task loss alone provides a short-term incentive to route many inputs through that module.
As a result, the routing objective must simultaneously produce sharp per-sample decisions while maintaining diversity in module usage at the population level.

For a routed layer \(\ell\), let \(E_{\ell,k}(x_i)\) denote the activation energy of the module \(k\) for input \(x_i\), as defined above. We convert the energy into a differentiable routing distribution as follows:

\begin{equation}
    q_{\ell,k}(x_i) = \frac{\exp(E_{\ell,k}(x_i)/\tau)} {\sum_{k'=1}^{K_\ell} \exp(E_{\ell,k'}(x_i)/\tau)},
    \label{eq:routing_distribution}
\end{equation}
where \(\tau > 0\) is a temperature parameter. Smaller values of \(\tau\) make the distribution closer to a winner-take-all decision, while larger values provide smoother gradients.

We minimize the entropy of the per-sample routing distribution to induce the first pressure, encouraging each input sample to select a distinct module:
\begin{equation}
    \mathcal{L}_{\mathrm{sample}}^{(\ell)} = \frac{1}{B} \sum_{i=1}^{B} H(q_\ell(x_i)) = -
    \frac{1}{B} \sum_{i=1}^{B} \sum_{k=1}^{K_\ell} q_{\ell,k}(x_i) \log q_{\ell,k}(x_i),
    \label{eq:sharp_loss}
\end{equation}
where \(B\) denotes the batch size.

We then add the second pressure that maximizes the entropy of the average usage distribution. This is a batch-level balance term under the assumption that, for a sufficiently large batch of inputs, module usage should be evenly distributed:
\begin{equation}
    \mathcal{L}_{\mathrm{bal}}^{(\ell)} = - H(\bar{q}_\ell) = \sum_{k=1}^{K_\ell} \bar{q}_{\ell,k} \log \bar{q}_{\ell,k}.
    \label{eq:balance_loss}
\end{equation}

This objective can be interpreted as encouraging high mutual information between inputs and module assignments, without using labels. The sharpness term reduces the conditional entropy of module assignment for each input, while the balance term raises the marginal entropy of module usage across inputs. Together, these two terms encourage confident, yet diverse, routing decisions. The task loss then determines which assignments are useful for prediction, allowing semantic specialization to emerge indirectly during training.

The final training objective is
\begin{equation}
    \mathcal{L} = \mathcal{L}_{\mathrm{task}} + \alpha \mathcal{L}_{\mathrm{route}}, \quad\mathcal{L}_{\mathrm{route}} = \mathcal{L}_{\mathrm{sample}} + \mathcal{L}_{\mathrm{bal}}
    \label{eq:full_loss}
\end{equation}
where \(\mathcal{L}_{\mathrm{task}}\) denotes the standard supervised loss function and $\alpha$ controls the strength of the routing objective $\mathcal{L}_{\mathrm{route}}$ during training.

\paragraph{Gaussian ring module masks for capacity competition.}

The default SCM formulation partitions a layer into $K$ contiguous blocks of equal width. This imposes a fixed-capacity prior in which each module receives exactly the same number of neurons. To relax this assumption, we replace hard block masks with soft Gaussian masks on a circular neuron axis. We decide to make the axis circular to avoid border effects: with 4 modules, the modules 0 and 3 are only in contact with one module, which would disadvantage them, where we want a fair competition between modules to avoid collapse. The ring simply allows each module to be connected to two modules.

Neuron $i$ and module $m$ are assigned positions $p_i=i/d_\ell$ and $c_m=m/K$ on the unit ring, and module memberships are computed as
\begin{equation}
    G_{i,m} = \operatorname{softmax}_{m} \left( -\frac{d_{\mathrm{ring}}(p_i,c_m)^2}{2\sigma_m^2 T_{\mathrm{gauss}}} \right),
\end{equation}

where $d_{\mathrm{ring}}$ denotes the periodic distance and $\sigma_m$ is a learnable module width. The widths are constrained by a fixed total budget, so increasing the capacity of one module reduces the capacity available to others. The routing energies are then computed as in the previous formulation, replacing the hard module mask $m_{\ell,k}$ with the soft membership vector $G_{:,k}$.

This modification does not change the SCM objective or the sparse top-1 routing rule. Instead, it replaces fixed modules of equal width with soft capacity-adaptive modules. As shown in Table~\ref{tab:fc2_module_count_ring}, the Gaussian ring does not substantially affect accuracy, while it consistently increases class-module mutual information. We therefore use SCM with Gaussian ring masks in the remaining experiments.

\begin{figure*}
\begin{center}
\includegraphics[width=0.9\linewidth]{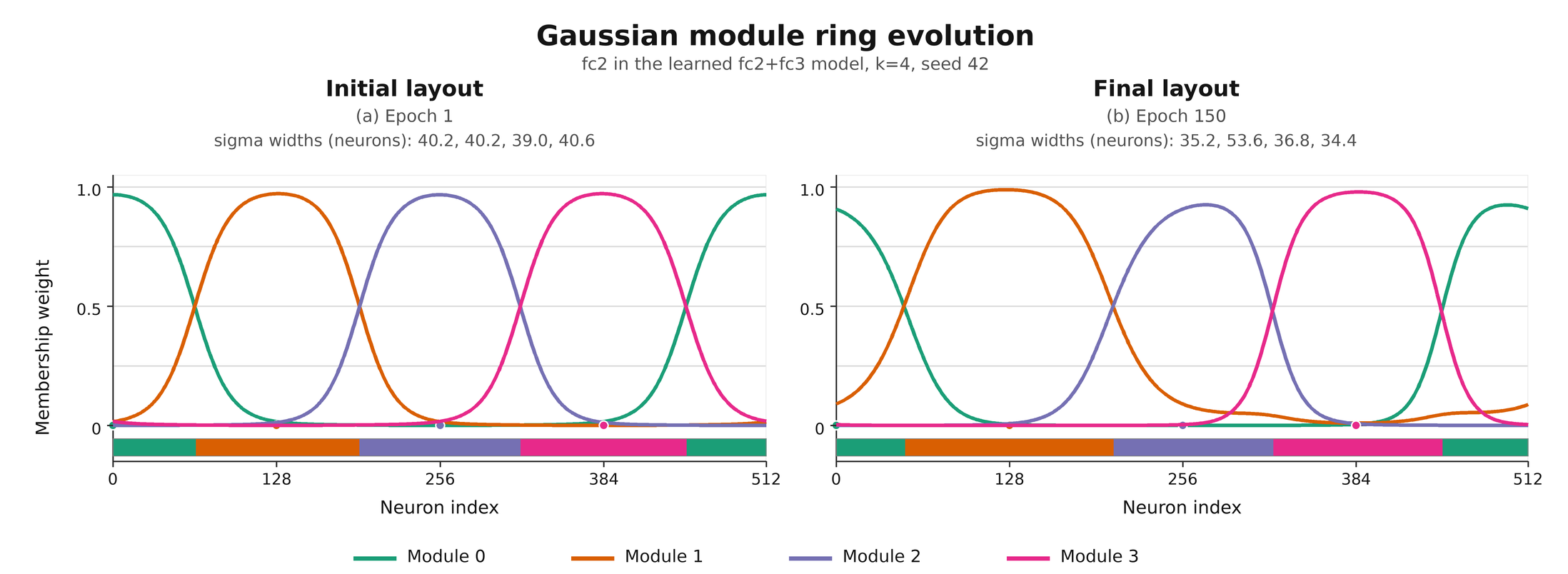}
\end{center}
   \caption{Illustration of the working of the Gaussian ring, with a run where the layer $fc2$ with 512 neurons is modularized into 4 modules. Left is the first epoch, and right is the last epoch. We can see the network learned to get a dominant module 1, without total collapse.}
\label{fig:gaussian_rings}
\end{figure*}

\subsection{Training and Inference}

The neural network is trained end-to-end using standard gradient-based training, with a hard top-1 mask applied during the forward, forcing the outputs of all the non-selected modules to be zero. Consequently, gradients obtained through the classification loss update only the selected module in that layer. The auxiliary routing loss is computed from the differentiable energy profile prior to the hard routing decision, allowing the model to reshape its activation energies during training to modify the preferred module assignment for a given input.

At inference time, we apply the same sparse top-1 routing rule as during training:
\begin{equation}
z_\ell(x)=\arg\max_k E_{\ell,k}(x),
\qquad
\tilde h_\ell(x)=m_{\ell,z_\ell(x)}\odot h_\ell(x).
\end{equation}


This consistency is important, as dense training relying on usage-based competition can yield class-structured energy profiles without producing modules that operate independently from the rest of the layer.  We therefore distinguish sparse competitive training from a dense control setting, as shown in Figure~\ref{fig:3_figures_training_evolution_horizontal}-A, in which all modules remain active during training and sparse top-1 routing is applied only at inference time. Task gradients flow only through the selected module, while the routing loss is computed from the pre-argmax distribution.

\subsection{Usage-Based Modularity Metrics}

Following prior work on clusterability-based modularity \cite{golechha2025studyingcrossclustermodularityneural}, we aim to quantify how routing induces an informative and balanced distribution of module usage. Ideally, routing should reflect specialized computation, such that different modules are selectively activated for different semantic subsets of data. In a classification task, subtasks are classifications of superclasses. Hence, a specialized module should be detected by a module only being activated when inputs corresponding to classes of a given superclass are present.


Then, we quantify class–module specialization for each routed layer \(\ell\) by computing the mutual information between the class variable \(C\) and the selected module index \(Z_\ell\):
\begin{equation}
    MI(C; Z_\ell) = \sum_{c,k} p(c,k) \log \frac{p(c,k)}{p(c)p(k)}.
\end{equation}

\section{Experiments}

\subsection{Setup}
\label{sec:setup}

\paragraph{Backbone.}
For comparability with the clusterability baseline, we use the same global architecture as \cite{golechha2025studyingcrossclustermodularityneural}: a small CNN stem followed by a bias-free fully connected head. After validating our method on CIFAR-10, we scaled the network to train from scratch on CIFAR-100 and ImageNet-100. Our \texttt{scaled\_cnn} maps $x \in \mathbb{R}^{3 \times 224 \times 224}$ through three Conv-BN-ReLU-MaxPool blocks with channels $32,64,128$, then adaptive-average-pools to $4 \times 4$, yielding a $2048$-dimensional feature vector. The head is $2048 \rightarrow 512 \rightarrow 512 \rightarrow 512 \rightarrow 100$, with ReLU activations after the first three fully connected layers with no bias terms. We denote the three hidden fully connected layers by  $fc1$,   $fc2$,  and  $fc3$,  where $fc1$ maps the convolutional feature vector to the first 512-dimensional hidden representation, while $fc2$ and $fc3$ are 512-to-512 hidden transformations. Routing is applied only to the selected layers among  $fc1$,   $fc2$,  and  $fc3$,  and the final $100$-way classifier remains dense.

\paragraph{Dataset and training.}

Our main experiments are conducted on ImageNet-100, a subset of 100 classes from the 1000-class ImageNet dataset. Instead of using a random subset, which could significantly affect the results, we use the standard CMC subset \cite{10.1007/978-3-030-58621-8_45}, which is designed to provide a more balanced selection of classes across diverse categories. All methods are evaluated with the same backbone and training protocol within each comparison. We report sweeps over routed layers ($fc2$,  $fc3$,  and $fc2+fc3$) and over the number of modules $K \in \{4,8,16\}$. To demonstrate that our results are not specific to ImageNet-100, we additionally report results on CIFAR-100.

Training images are processed with ImageNet normalization. Validation images are resized to $256$, center-cropped to $224$, and normalized in the same manner. Models are trained for 150 epochs with Adam, a batch size of 128, a learning rate of $10^{-3}$. Unless stated otherwise, we use a constant $\alpha=0.1$.

We compare Sparse Competitive Modules against three baselines. The vanilla baseline is trained with the standard classification loss and no modular objective. The clustered baseline follows the clusterability approach of \cite{golechha2025studyingcrossclustermodularityneural}, which encourages weights to concentrate within predefined module groups. Unless otherwise stated, SCM refers to sparse competitive modules with Gaussian ring module masks. Finally, to show that with our method we in fact learn in a similar way as Mixture-of-Experts, but without a learned router, we add a learned router baseline. This baseline uses the same backbone, routed layers, module masks, and number of modules as SCM, but replaces the energy-based routing distribution with a learned linear router $q_\ell(x)=\mathrm{softmax}(W_\ell h_\ell(x)/\tau)$. During training and evaluation, the selected module is the top-1 router prediction, and all non-selected modules are masked out, as in SCM. The router is trained jointly with the classification objective and standard load-balancing auxiliary losses to avoid collapse.
Routing probabilities are computed from module activation energies using temperature $\tau=1.0$. 

\begin{figure}[!t]
\begin{center}
\includegraphics[width=0.8\linewidth]{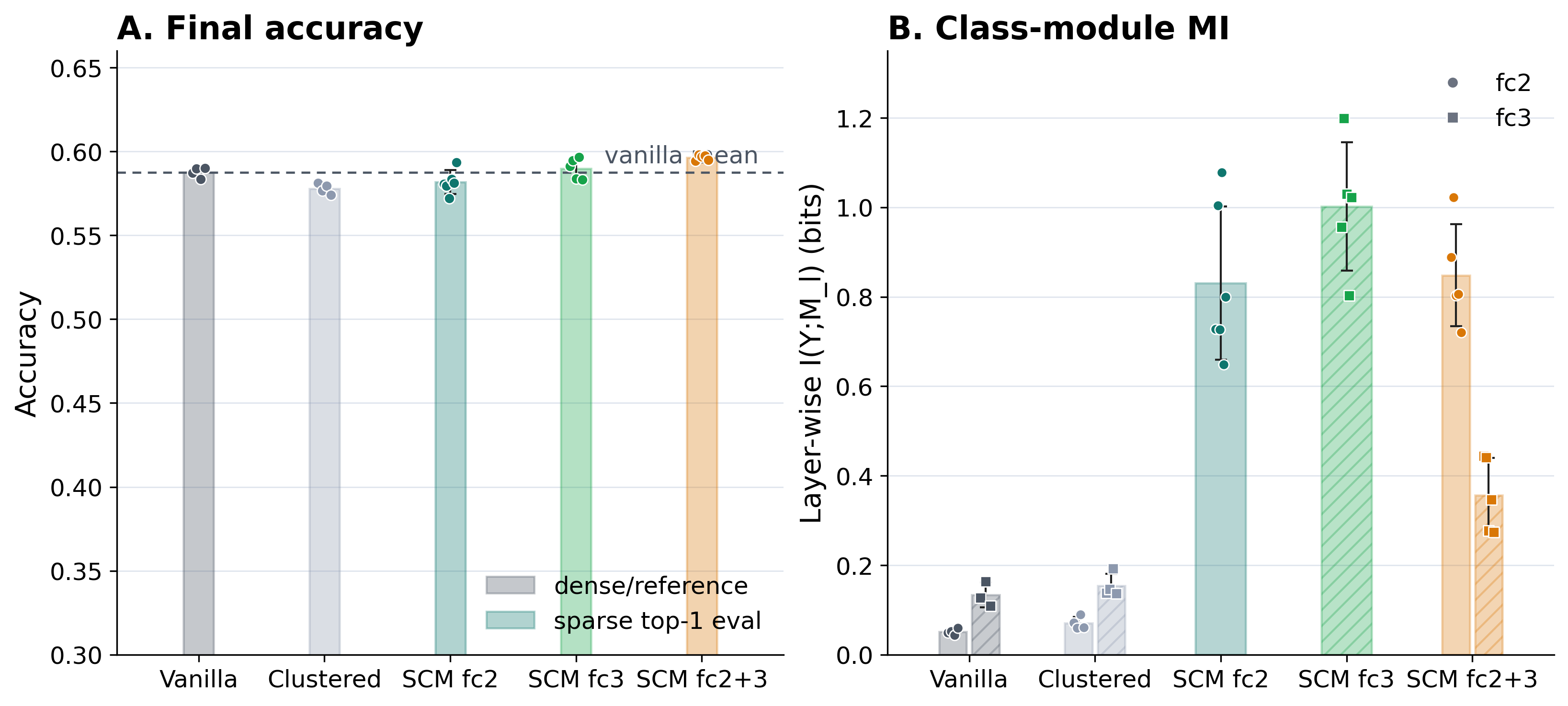}
\end{center}
   \caption{Measures of accuracy and MI of the same CNN model after training on ImageNet-100 CMC. A: The accuracies of our method, clustered baseline from \cite{golechha2025studyingcrossclustermodularityneural} and without modularization. Here, the accuracy of our method corresponds to the top-1 accuracy while other methods are evaluated on dense accuracies. B: the MI between the favored module and the input class.}
\label{fig:deployable_specialization_hierarchy}
\end{figure}

\paragraph{Evaluation.} 
For our method, training and inference use the same sparse top-1 routing rule: only the selected module remains active in a routed layer. We also include a dense-control variant, where the routing objective is used during training but all modules remain active in the forward pass. This separates semantic specialization of the routing scores from sparse deployment of the selected modules.

We report classification accuracy and class-module mutual information $MI(Y;M_\ell)$, where $M_\ell$ is the selected module in layer $\ell$. To detect collapse and measure the evenness of usage, we also report the effective number of used modules:
\begin{equation}
    N_{\mathrm{eff}} = \exp(H(M_\ell)).
\end{equation}

For semantic analysis, we measure whether classes from the same coarse ImageNet group are assigned to the same module more often than expected by chance. In the module-count sweep, we additionally measure whether higher-\(K\) partitions refine lower-\(K\) partitions using pair retention and weighted child purity.

Finally, to test robustness to the choice of classes, we repeat the main ImageNet-100 experiment on two random class subsets. We report mean and standard deviation across matched seeds when available.

\subsection{Results}

\paragraph{Our method maintains near-baseline accuracy with module specialization}
Table~\ref{tab:main_fc2_imagenet_cifar} reports the main ImageNet-100 CMC comparison for routing in $fc2$ with $k=4$ modules. These results are also shown in Figure~\ref{fig:deployable_specialization_hierarchy}. Our method obtains $57.84 \pm 0.59\%$, matching the clustered baseline and remaining within roughly one percentage point of the vanilla model. Thus, SCM does not bring accuracy degradation. We can observe the same on CIFAR-100.

The main difference is not accuracy but the information carried by module usage. Vanilla and clustered models have low class-module mutual information at  $fc2$,  with $MI(Y;M_{fc2})=0.036 \pm 0.006$ and $0.049 \pm 0.012$, respectively. In contrast, our method reaches $0.649 \pm 0.128$, an increase of approximately $18 \times$ over vanilla and $13 \times$ over the clustered baseline. This shows that the selected module is no longer an arbitrary subdivision of the layer: it becomes a class-informative routing variable.

\begin{table}[t]
  \centering
  \small
  \caption{
Main comparison on ImageNet-100 CMC and CIFAR-100 with
\texttt{fc2} routed. Accuracy uses each method's intended inference
mode; MI is computed from the selected/highest-energy module.
}
  \label{tab:main_fc2_imagenet_cifar}
  \begin{tabular}{llcc}
    \toprule
    Dataset & Method & Acc. (\%) & $\mathrm{MI}(Y;M_{\texttt{fc2}})$ \\
    \midrule
    ImageNet-100 CMC
      & Vanilla & $58.77 \pm 0.38$ & $0.036 \pm 0.006$ \\
    ImageNet-100 CMC
      & Clustered baseline & $57.67 \pm 0.26$ & $0.049 \pm 0.012$ \\
    ImageNet-100 CMC
      & SCM & $57.84 \pm 0.59$ & $0.649 \pm 0.128$ \\
    \midrule
    CIFAR-100
      & Vanilla & $47.76 \pm 2.10$ & $0.114 \pm 0.049$ \\
    CIFAR-100
      & Clustered baseline & $49.00 \pm 1.10$ & $0.139 \pm 0.037$ \\
    CIFAR-100
      & SCM & $48.94 \pm 2.40$ & $0.493 \pm 0.194$ \\
    \bottomrule
  \end{tabular}
\end{table}

To test whether the CMC split contains an unusually favorable semantic structure, we repeat the experiment on two random ImageNet-100 label subsets. The results in the table of section C of the Appendix show the same qualitative pattern. Thus, the specialization effect is not specific to the CMC label subset: across different class selections, SCM preserves baseline-level accuracy while making module assignments more predictive of class identity.

\paragraph{Increasing the number of modules gives finer specialization, Gaussian ring too.}

Table~\ref{tab:fc2_module_count_ring} isolates the effect of the Gaussian ring masks. The Gaussian ring does not substantially change the accuracy: across $K \in \{4,8,16\}$, the difference relative to SCM without the ring is small. However, it consistently increases $MI(Y;M_{\mathrm{fc2}})$, from $0.409$ to $0.638$ at $K=4$, from $0.730$ to $0.891$ at $K=8$, and from $1.150$ to $1.297$ at $K=16$. Hence, we use SCM with Gaussian-ring competition in the remaining experiments, not for the accuracy, but as a mechanism that strengthens modular specialization.

The learned MoE baseline on Table~\ref{tab:fc2_module_count_ring} also shows that we can achieve the accuracy and specialization of a learned router without explicitly using one. At $K=4$, it reaches a similar accuracy and mutual information to SCM, but as the number of modules increases, the learned MoE baseline loses in accuracy: $2.3$ and $5.8$ points in accuracy compared to our method while having a lower MI. The module usage also shows a worse balance overall. Our method is hence able to reproduce (or even surpass in more high pressure settings) a learned router, only with new training dynamics.

\begin{figure}[!t]
\begin{center}
\includegraphics[width=\linewidth]{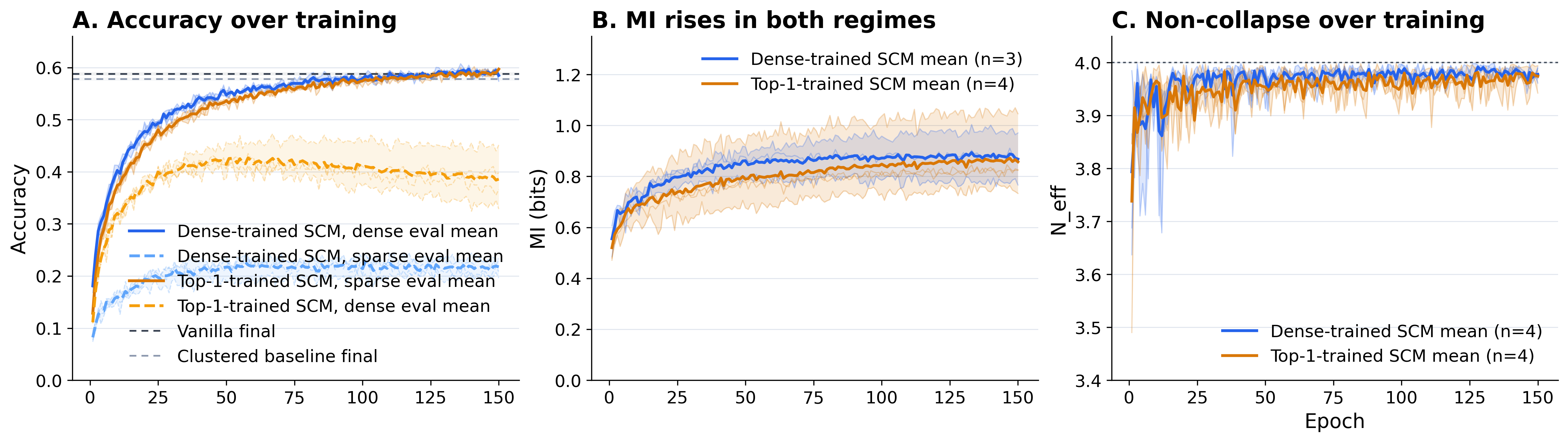}
\end{center}
   \caption{Evolution of metrics during training on ImageNet-100 CMC. A. The top-1 accuracies of our method (SCM) compared to no clustering (Vanilla) and the method from \cite{filan2021clusterabilityneuralnetworks} (Clustered); in dotted lines are the opposite evaluations compared to training: the dense training is on sparse eval, and the sparse training is on dense eval. Both regimes perform better when evaluated like training, but sparse training is better on dense than dense on sparse eval. B. The evolution of MI between our method trained in sparse and dense training. C. The evolution of the effective number of modules at the end of training. The trainings on our method converge on $N_{eff}=4$ meaning all modules are used evenly.}
\label{fig:3_figures_training_evolution_horizontal}
\end{figure}

\paragraph{Sparse training is necessary for sparse deployment}
Figures~\ref{fig:3_figures_training_evolution_horizontal}-A and B separate two notions of modularity. A dense-control model can learn class-structured routing scores while all modules remain active in the forward pass. However, because every training example can rely on all modules, the selected module is not forced to be individually sufficient. Applying top-1 routing only at evaluation therefore produces modules that are semantically recognizable but not reliably deployable. In contrast, SCM trains and evaluates with the same hard top-1 mask. The selected module receives the task gradient for that example, while non-selected modules do not. This forces each routed module to support the computation assigned to it. The ablation therefore shows that semantic specialization of routing scores is not enough: sparse competitive training is required to obtain modules that remain functional under sparse inference.

\paragraph{Modules recover coarse semantic structure: the emergence of a hierarchy of classes}
For each class, we define its dominant module as the module selected most often on validation examples, and compare the resulting partition with coarse ImageNet groups, using these groups only for evaluation. Classes from the same coarse group are routed to the same module roughly \(3\times\) more often than chance, suggesting that the modules capture semantic structure rather than arbitrary class identities. We further test whether this structure forms a hierarchy by training independent $fc2$ models with \(K \in \{4,8,16\}\) modules. SCM shows clear coarse-to-fine consistency: pair retention is \(28.32\%\) for \(4\!\to\!8\) and \(27.15\%\) for \(8\!\to\!16\), compared with random baselines of \(13.85\pm0.86\%\) and \(6.40\pm0.89\%\). Weighted child purity is also high, at \(69.0\%\) and \(72.0\%\). Thus, increasing the number of modules refines broad semantic specialists into finer ones, rather than producing unrelated partitions.

\begin{table}[t]
\centering
\small
\caption{
Module-count sweep at \texttt{fc2} on ImageNet-100 CMC.
MI is measured in nats; $MI/\ln K$ is normalized MI.
Clustered is evaluated densely; SCM and Learned-MoE use sparse top-1 inference.
}
\label{tab:fc2_module_count_ring}
\setlength{\tabcolsep}{4.2pt}
\begin{tabular}{@{}llcccc@{}}
\toprule
$K$ & Method & Acc. (\%) & $MI(Y;M_{\mathrm{fc2}})$ & $MI/\ln K$ & $N_{\mathrm{eff}}$ \\
\midrule
4 & Clustered baseline
  & $57.92 \pm 0.22$ & $0.048 \pm 0.015$ & $0.035 \pm 0.011$ & $3.83 \pm 0.02$ \\
4 & Learned-MoE
  & $57.60 \pm 0.70$ & $0.50 \pm 0.10$ & $0.361 \pm 0.072$ & $3.25 \pm 0.01$ \\
4 & SCM w/o Gaussian ring
  & $57.53 \pm 0.81$ & $0.409 \pm 0.091$ & $0.295 \pm 0.066$ & $3.96 \pm 0.02$ \\
4 & SCM + Gaussian ring
  & $57.92 \pm 0.23$ & $0.638 \pm 0.111$ & $0.460 \pm 0.080$ & $3.97 \pm 0.01$ \\

\midrule
8 & Clustered baseline
  & $58.74 \pm 0.44$ & $0.062 \pm 0.012$ & $0.030 \pm 0.006$ & $7.83 \pm 0.03$ \\
8 & Learned-MoE
  & $55.50 \pm 1.00$ & $0.78 \pm 0.12$ & $0.375 \pm 0.058$ & $6.45 \pm 0.03$  \\
8 & SCM w/o Gaussian ring
  & $57.61 \pm 0.46$ & $0.730 \pm 0.039$ & $0.351 \pm 0.019$ & $7.85 \pm 0.05$ \\
8 & SCM + Gaussian ring
  & $57.79 \pm 0.22$ & $0.891 \pm 0.050$ & $0.428 \pm 0.024$ & $7.83 \pm 0.08$ \\

\midrule
16 & Clustered baseline
  & $57.56 \pm 0.87$ & $0.113 \pm 0.019$ & $0.041 \pm 0.007$ & $15.09 \pm 0.08$ \\
16 & Learned-MoE
  & $51.00 \pm 1.50$ & $0.90 \pm 0.15$ & $0.325 \pm 0.054$ & $14.34 \pm 0.09$  \\
16 & SCM w/o Gaussian ring
  & $55.77 \pm 0.73$ & $1.150 \pm 0.030$ & $0.415 \pm 0.011$ & $15.31 \pm 0.12$ \\
16 & SCM + Gaussian ring
  & $56.81 \pm 0.38$ & $1.297 \pm 0.026$ & $0.468 \pm 0.009$ & $15.52 \pm 0.10$ \\
\bottomrule
\end{tabular}
\end{table}

\subsection{More analysis}

\paragraph{Causal specialization tests}

\begin{table}[t]
\centering
\small
\caption{
Causal specialization diagnostic on ImageNet-100 with four modules at
\texttt{$fc2$} and \texttt{$fc3$}. Classes are assigned to their dominant module;
we then ablate each module and measure the accuracy drop on its own classes
and on all other classes. Drops are percentage points. Full per-module results
are given in Appendix B.
}
\label{tab:causal_specialization}
\setlength{\tabcolsep}{4.2pt}
\begin{tabular}{llrrrr}
\toprule
Model & Layer & Owned drop & Other drop & Gap & Ownership $N_{\mathrm{eff}}$ \\
\midrule
SCM & \texttt{$fc2$} & $22.38$ & $11.39$ & $10.99$ & $3.99$ \\
SCM & \texttt{$fc3$} & $27.32$ & $9.17$  & $18.15$ & $3.99$ \\
\midrule
Clustered & \texttt{$fc2$} & $37.40$ & $15.50$ & $21.90$ & $2.48$ \\
Clustered & \texttt{$fc3$} & $24.48$ & $10.59$ & $13.89$ & $1.71$ \\
\midrule
Vanilla & \texttt{$fc2$} & $32.22$ & $21.64$ & $10.58$ & $3.39$ \\
Vanilla & \texttt{$fc3$} & $19.90$ & $13.13$ & $6.77$  & $3.32$ \\
\bottomrule
\end{tabular}
\end{table}

Finally, we test whether class-module associations are causally relevant rather than merely correlational. For each class, we identify its dominant module, ablate that module at test time, and compare the accuracy drop on classes owned by the ablated module with the drop on all other classes. Table~\ref{tab:causal_specialization} shows that SCM has positive causal gaps in both routed layers: $10.99$ percentage points in \texttt{$fc2$} and $18.15$ in \texttt{$fc3$}. Importantly, SCM maintains uniform ownership across the four modules.

The clustered baseline also exhibits causal gaps, but its ownership is highly imbalanced, especially in \texttt{$fc3$}, where the effective number of class-owning modules drops to $1.71$. Thus, SCM does not merely increase class-module mutual information; it induces a balanced decomposition whose modules are functionally relevant for their assigned classes. We report these results in Appendix B.

\paragraph{Routing deeper layers improves the accuracy--specialization tradeoff.}

Next, we ask where SCM should be applied. Table~\ref{tab:layer_placement_sweep} reports a layer-placement sweep on ImageNet-100 CMC using the same \texttt{scaled\_cnn}, and \(K=4\) modules. Routing in $fc1$ already induces important class-module specialization, but it comes with a $3.04\%$ accuracy drop relative to the matched dense baseline. Hence, early fully connected features are still too general: forcing a sparse module choice at this stage removes useful shared computation.

Routing later layers gives a better tradeoff. In $fc2$, sparse top-1 routing slightly exceeds the matched baseline accuracy, $58.06\%$ versus $57.92\%$, while increasing class-module mutual information from $0.0476$ to $0.5049$. Routing only \texttt{$fc3$} gives the best single-layer specialization, with $MI(Y;M)=0.6628$, and also improves accuracy by $0.32$ percentage points. Routing both $fc2$ and $fc3$ gives the best overall result, with $59.42\%$ top-1 accuracy and high MI in both layers. This shows that modularity works best in later layers, where features are already semantically organized enough for module selection to be class-informative.

\begin{table}[t]
  \centering
  \small
  \setlength{\tabcolsep}{4.5pt}
  \caption{
  Layer-placement sweep on ImageNet-100 CMC. All rows use \texttt{scaled\_cnn}, seed 42, 150 epochs, and \(K=4\) modules. Accuracy values are percentages.
  }
  \label{tab:layer_placement_sweep}
  \begin{tabular}{lccccc}
    \toprule
    Routed layers & Baseline acc. & Ours dense acc. & Ours top-1 acc. & Baseline MI & Ours MI \\
    \midrule
    \texttt{$fc1$}
    & 58.80 & 46.98 & 57.16 & 0.0179 & 0.4059 \\
    \texttt{$fc2$}
    & 57.92 & 53.54 & 58.06 & 0.0476 & 0.5049 \\
    \texttt{$fc3$}
    & 58.80 & 55.90 & 59.12 & 0.0763 & 0.6628 \\
    \texttt{$fc2+fc3$}
    & 58.12 & 36.24 & 59.42 & 0.049 / 0.095 & 0.615 / 0.307 \\
    \bottomrule
  \end{tabular}
\end{table}

\begin{table}[t]
  \centering
  \small
  \setlength{\tabcolsep}{6pt}
  \caption{Hierarchy metrics for the \texttt{$fc2$} module-count sweep on ImageNet-100 CMC.}
  \label{tab:hierarchy_retention_purity}
  \begin{tabular}{llccc}
    \toprule
    Method & Transition & Pair retention & Random baseline & Weighted child purity \\
    \midrule
    Ours & $4 \to 8$ & $28.32$ & $13.85 \pm 0.86$ & $69.00$ \\
    Ours & $8 \to 16$ & $27.15$ & $6.40 \pm 0.89$ & $72.00$ \\
    Clustered & $4 \to 8$ & $21.70$ & $23.72 \pm 1.55$ & $66.00$ \\
    Clustered & $8 \to 16$ & $11.33$ & $9.62 \pm 0.83$ & $48.00$ \\
    \bottomrule
  \end{tabular}
\end{table}

\begin{figure*}
\begin{center}
\includegraphics[width=\linewidth]{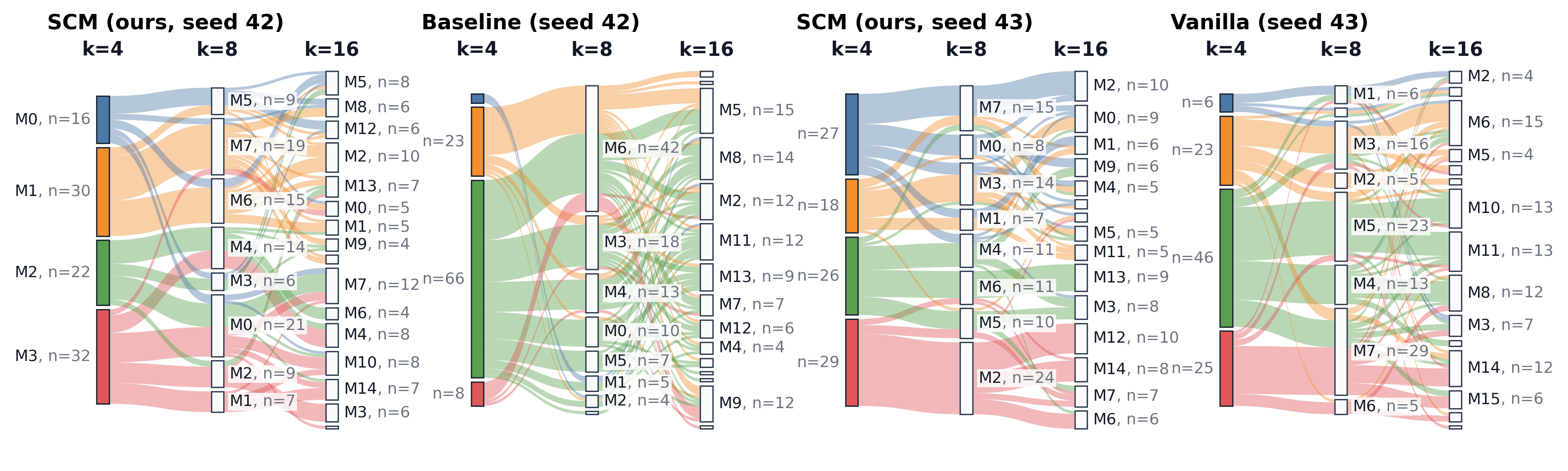}
\end{center}
   \caption{
Alluvial visualization of class assignments across independently trained $fc2$ runs with $K=4,8,16$ modules. Each block is a learned module and each flow tracks ImageNet-100 classes across module-count resolutions. We compare the cleaner refinement structure obtained by our method against the two alternative flows shown in the figure. To rule out this structure being a seed artifact, we also show two independent seeds of our method, both of which exhibit similar coarse-to-fine class flows.
}
\label{fig:hierarchy_alluvial_fc2}
\end{figure*}

\paragraph{Higher-resolution module partitions refine lower-resolution ones}
We next ask whether increasing $k$ in completely unrelated runs produces unrelated partitions or a hierarchy of refinements that is compatible across seeds and module numbers: if modules meaningfully partition data on consistent granularity. Table~\ref{tab:hierarchy_retention_purity} reports pair retention and weighted child purity for independently trained $K=4,8,16$ runs.
We compare two independently trained partitions using two metrics. \emph{Pair retention} is the fraction of class pairs assigned to the same module at lower $k$ that remain co-assigned at higher $k$. We compare it to a random baseline obtained by permuting the higher-$k$ assignments while preserving module sizes. \emph{Weighted child purity} measures whether each higher-$k$ module mostly comes from a single lower-$k$ parent, weighted by the number of classes in the child module.
Table~\ref{tab:hierarchy_retention_purity} shows that SCM obtains a clear refinement structure. SCM has pair retention well above the random baseline for both $4{\to}8$ and $8{\to}16$ transitions. These fare better than the clustered baseline, being closer or even below their random baseline, and lower than our method. SCM also obtains higher weighted child purity than the clustered baseline.
The alluvial visualization in Figure~\ref{fig:hierarchy_alluvial_fc2} illustrates the same pattern: SCM modules at higher resolution behave like semantic subdivisions of lower-resolution modules, whereas clustered partitions are less stable across independently trained resolutions.

\begin{table*}[t]
\centering
\small

\begin{minipage}[t]{0.48\textwidth}
\centering
\textbf{(a) Architecture transfer}\\[3pt]
\setlength{\tabcolsep}{5pt}
\begin{tabular}{@{}llcc@{}}
\toprule
Backbone & Layer & Vanilla acc. & SCM acc. / MI \\
\midrule
ResNet-18 & \texttt{fc2} & 76.12 & 76.88 / 1.065 \\
ViT-Tiny  & \texttt{fc2} & 80.66 & 80.26 / 1.117 \\
ViT-Tiny  & \texttt{fc3} & 80.30 & 80.70 / 1.130 \\
\bottomrule
\end{tabular}
\end{minipage}
\hfill
\begin{minipage}[t]{0.48\textwidth}
\centering
\textbf{(b) ResNet-18 hierarchical refinement}\\[3pt]
\setlength{\tabcolsep}{5pt}
\begin{tabular}{@{}lccc@{}}
\toprule
Transition & Pair ret. & Random & Child purity \\
\midrule
$4\!\to\!8$  & 28\% & 13\% & 70\% \\
$8\!\to\!16$ & 30\% & 7\%  & 72\% \\
\bottomrule
\end{tabular}
\end{minipage}

\caption{
Transfer beyond scaled\_cnn on ImageNet-100 CMC.
(a) Accuracy and class-module MI for ResNet-18 and ViT-Tiny with $K=4$;
SCM is evaluated with sparse top-1 inference.
(b) Coarse-to-fine refinement across independently trained ResNet-18
module-count configurations.
}
\label{tab:architecture_transfer}
\end{table*}

\paragraph{Transfer across architectures.}
We evaluate our method on more limited settings on ResNet-18 and ViT-Tiny, with results on Table~\ref{tab:architecture_transfer}. On ResNet-18, SCM reaches $76.88\%$ accuracy with $MI(Y;M)=1.065$, compared with $76.12\%$ accuracy for the vanilla model. On ViT-Tiny, SCM increases MI from $0.074$ to $1.117$ at \texttt{fc2} with only a $0.40$-point accuracy cost, while at \texttt{fc3} it improves accuracy ($80.70\%$ vs.\ $80.30\%$) and obtains strong MIs. Table~\ref{tab:architecture_transfer} also show that the coarse-to-fine organization also transfer to ResNet-18, as pair retention is $28\%$ and $30\%$ for $4\!\to\!8$ and $8\!\to\!16$, respectively, well above random baselines of $13\%$ and $7\%$, with child purities of $70\%$ and $72\%$. These results show that SCM transfers across convolutional, residual, and Transformer architectures.

\section{Conclusion}
In this work we proposed a method that yields modules in neural networks through competition. The competition is mainly carried through usage-based competition with Sparse Competitive Modules allowing for a modularization that produces meaningfully specialized modules. Our second, complementary way to further increase specialization of modules is through capacity-based competition with Gaussian ring ownership of neurons, where the width of modules is learned. These two ways, as we showed, allow for modules that conserve the accuracy of a non-modularized network, while meaningfully gathering the data into groups at varying granularity, adapting to its number of modules. This work demonstrates a new way to modularize networks: through the loss only, via information theoretical metrics. This approach further manifests that the internal organization does not need to be monitored, but given the right incentives, it can appear spontaneously.

\section{Limitations and future works}

Our experiments are intentionally conducted in a controlled setting, which makes module usage, specialization, and ablation easy to measure. A natural next step is to apply the same objective to larger backbones, larger vision datasets, and models with richer internal structure, language models, and vision-language models.

A second limitation is that only a small number of layers are modularized. Most experiments route one layer, and our multi-layer results are limited to 1 to 3 layers. This is enough to show that routing becomes class-informative and remains usable under sparse inference, but not enough to claim full-network modular computation or complete circuit recovery. The main interpretability promise of modular training is to obtain whole \emph{circuits}: connected chains of modules that support particular computations across depth. Our preliminary multi-layer experiment in Appendix A, where $fc2$ is routed into four modules and $fc3$ into eight modules, suggests that this may occur naturally. This suggests that sparse competitive routing may already bias the network toward structured module-to-module pathways.

\paragraph{Aknowledgements.}

This work was funded by the HOLIGRAIL project (ANR-23-PEIA-0010). Computing and storage resources were provided by GENCI at IDRIS under grant 2026-AD011016984 on the H100 partition of the Jean Zay supercomputer.

\bibliography{egbib}
\end{document}


\maketitle

This supplementary material is used as the appendix of the paper Sparse Competition during Training for the Emergence of Specialized Modules. Unless otherwise stated, all additional analyzes use the same ImageNet-100
\texttt{scaled\_cnn} setup as in the main paper. We report two complementary
diagnostics. First, we visualize whether modular assignments across two routed
layers form a coherent cross-layer structure. Second, we provide the full per-module breakdown of the causal specialization test summarized in the main paper.

\begin{figure*}
\begin{center}
\includegraphics[width=0.8\linewidth]{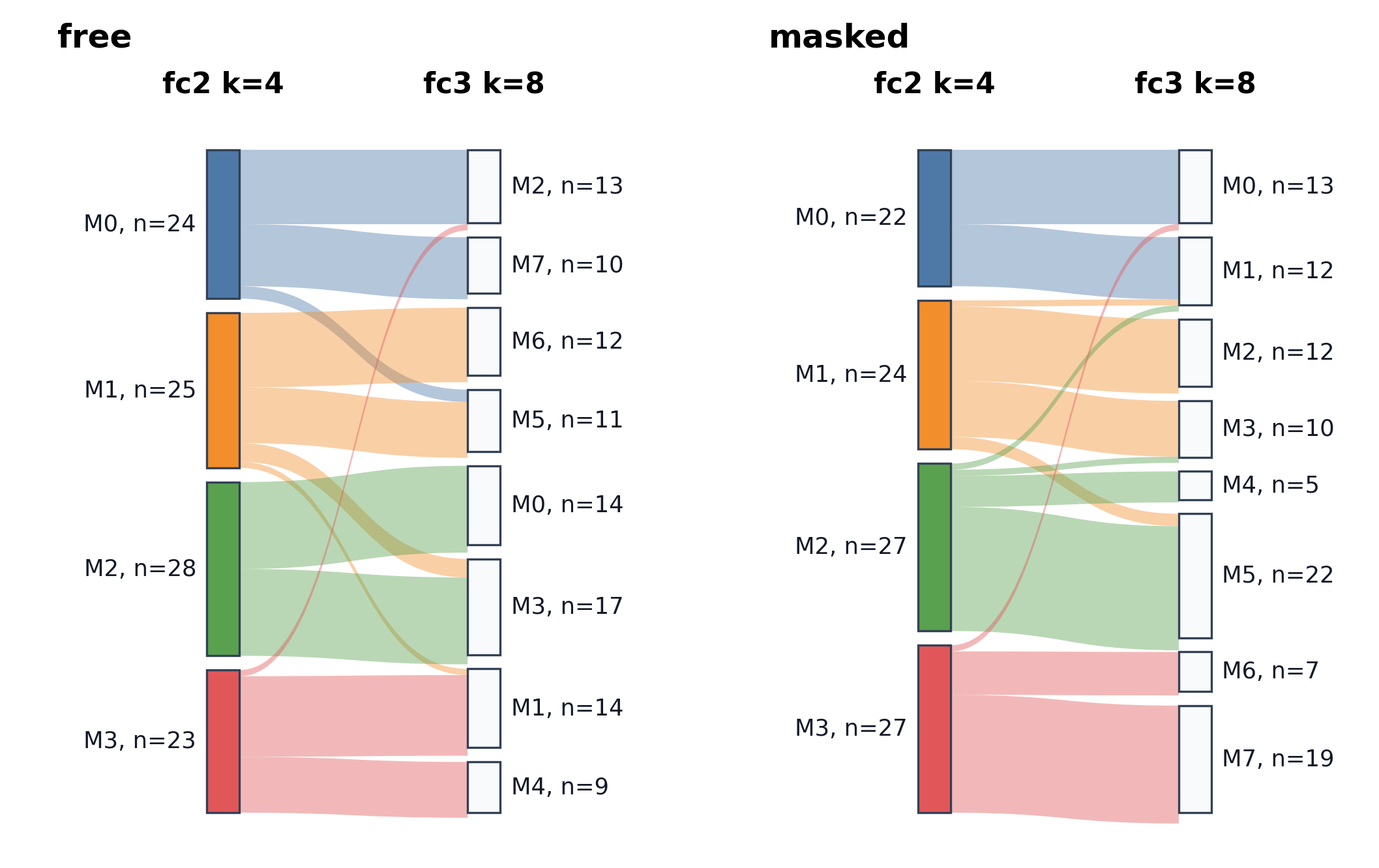}
\end{center}
\caption{
Alluvial visualization of shared dominant classes between a routed \texttt{fc2}
layer with $K=4$ modules and a routed \texttt{fc3} layer with $K=8$ modules.
We compare an unconstrained model with a hierarchy-constrained variant in which
each \texttt{fc2} module can connect only to two designated \texttt{fc3} modules.
The two diagrams are qualitatively similar, indicating that cross-layer modular
structure can arise without explicitly enforcing a hierarchical routing prior.
}
\label{fig:appendix_alluvial}
\end{figure*}

\section*{A. Cross-layer modular structure}

When more than one layer is routed, a natural question is whether modules at different layers form coherent multi-layer circuits, or whether their assignments are independent across layers. To investigate this, we compare two variants with a routed \texttt{fc2} layer using $K=4$ modules and a routed \texttt{fc3} layer using $K=8$ modules. In the unconstrained variant, no explicit relation is imposed between the modules of the two layers. In the constrained variant, we impose a hard hierarchical connectivity prior: each \texttt{fc2} module is allowed to feed only two designated \texttt{fc3} modules.

Figure~\ref{fig:appendix_alluvial} shows the resulting alluvial diagrams. Each flow represents classes that share the same dominant module at \texttt{fc2} and \texttt{fc3}. The constrained and unconstrained variants produce qualitatively similar flows, suggesting that cross-layer modular organization can arise without explicitly imposing a parent--child module hierarchy. This diagnostic is qualitative, but it supports the view that SCM can align modular assignments across layers through the training objective itself, rather than through a hand-designed connectivity prior.

\section*{B. Per-module causal specialization}

The main paper reports an aggregate causal specialization diagnostic. Here we provide the full per-module breakdown. For each trained model and each routed layer, we first assign every class to its dominant module. We then ablate one module at test time and measure the accuracy drop on the classes owned by the ablated module and on all remaining classes. A positive gap indicates that the ablated module is more important for the classes assigned to it than for other classes.

Table~\ref{tab:appendix_causal_per_module} shows that SCM has a positive causal gap for every module at both \texttt{fc2} and \texttt{fc3}. Moreover, SCM assigns classes nearly uniformly across modules: each module owns between $23$ and $28$ classes. The clustered baseline also exhibits positive causal gaps, but with substantial ownership imbalance. In particular, at \texttt{fc3}, one clustered module owns $86$ out of $100$ classes, while the other three modules own only $4$, $8$, and $2$ classes. Vanilla networks also show positive gaps, but the effect is weaker at \texttt{fc3}. These results support the interpretation that SCM induces balanced, causally relevant specialization rather than only class-correlated routing assignments.

\begin{table*}
\centering
\caption{
Per-module causal specialization on ImageNet-100 using scaled\_cnn, seed 42, and four-module \texttt{fc2}+\texttt{fc3} runs. Classes are assigned to their dominant module, defined as the module selected most often for that class. Each row ablates one module at test time. \emph{Owned classes} counts the classes assigned to the ablated module. \emph{Owned drop} and \emph{Other drop} are the corresponding accuracy decreases, in percentage points, on owned classes and on all remaining classes. \emph{Gap} is the difference between these two drops; positive values indicate class-specific causal specialization.
}
\vspace{0.5em}
\label{tab:appendix_causal_per_module}
\setlength{\tabcolsep}{4.2pt}
\begin{tabular}{llrrrrr}
\toprule
Model & Layer & Module & Owned classes & Owned drop & Other drop & Gap \\
\midrule
SCM & \texttt{fc2} & 1 & 25 & $20.96$ & $8.24$  & $12.72$ \\
SCM & \texttt{fc2} & 2 & 23 & $23.30$ & $13.22$ & $10.08$ \\
SCM & \texttt{fc2} & 3 & 24 & $20.00$ & $14.39$ & $5.61$ \\
SCM & \texttt{fc2} & 4 & 28 & $24.93$ & $9.53$  & $15.40$ \\
SCM & \texttt{fc3} & 1 & 24 & $23.83$ & $11.95$ & $11.89$ \\
SCM & \texttt{fc3} & 2 & 25 & $25.04$ & $9.65$  & $15.39$ \\
SCM & \texttt{fc3} & 3 & 23 & $25.04$ & $9.38$  & $15.67$ \\
SCM & \texttt{fc3} & 4 & 28 & $34.21$ & $5.53$  & $28.69$ \\
\midrule
Clustered baseline & \texttt{fc2} & 1 & 12 & $51.83$ & $15.89$ & $35.95$ \\
Clustered baseline & \texttt{fc2} & 2 & 16 & $46.75$ & $18.12$ & $28.63$ \\
Clustered baseline & \texttt{fc2} & 3 & 3  & $67.33$ & $12.31$ & $55.02$ \\
Clustered baseline & \texttt{fc2} & 4 & 69 & $31.42$ & $17.29$ & $14.13$ \\
Clustered baseline & \texttt{fc3} & 1 & 4  & $38.00$ & $12.23$ & $25.77$ \\
Clustered baseline & \texttt{fc3} & 2 & 8  & $50.25$ & $12.43$ & $37.82$ \\
Clustered baseline & \texttt{fc3} & 3 & 2  & $44.00$ & $8.08$  & $35.92$ \\
Clustered baseline & \texttt{fc3} & 4 & 86 & $21.00$ & $4.86$  & $16.14$ \\
\midrule
Vanilla & \texttt{fc2} & 1 & 39 & $25.03$ & $17.74$ & $7.29$ \\
Vanilla & \texttt{fc2} & 2 & 38 & $38.74$ & $26.87$ & $11.87$ \\
Vanilla & \texttt{fc2} & 3 & 15 & $34.80$ & $24.78$ & $10.02$ \\
Vanilla & \texttt{fc2} & 4 & 8  & $31.50$ & $17.80$ & $13.70$ \\
Vanilla & \texttt{fc3} & 1 & 34 & $20.06$ & $13.97$ & $6.09$ \\
Vanilla & \texttt{fc3} & 2 & 10 & $21.20$ & $15.42$ & $5.78$ \\
Vanilla & \texttt{fc3} & 3 & 11 & $16.18$ & $11.19$ & $4.99$ \\
Vanilla & \texttt{fc3} & 4 & 45 & $20.40$ & $11.49$ & $8.91$ \\
\bottomrule
\end{tabular}
\end{table*}

\section*{C. Robustness of the ImageNet-100 CMC results}

To test whether the CMC ImageNet-100 split contains an unusually favorable semantic
structure, we repeat the main routed-\texttt{fc2} experiment on two random 100-class
subsets. Table~\ref{tab:random_subset_results} shows that SCM preserves baseline-level accuracy
while substantially increasing class--module mutual information across both subsets.

\begin{table}
\centering
\small
\setlength{\tabcolsep}{6pt}
\caption{
Results across two random ImageNet-100 label subsets. Sparse top-1 SCM preserves baseline-level accuracy while substantially increasing class-module mutual information. Bold denotes the best value within each metric group.
}
\vspace{0.5em}
\label{tab:random_subset_results}
\begin{tabular}{lcccccc}
\toprule
\multirow{2}{*}{Subset} 
& \multicolumn{4}{c}{Accuracy} 
& \multicolumn{2}{c}{MI$(\mathrm{class}; \mathrm{module})$} \\
\cmidrule(lr){2-5}
\cmidrule(lr){6-7}
& Vanilla & Clustered & SCM dense & SCM top-1 & SCM & Clustered \\
\midrule
Label seed 42 & 0.5528 & 0.5528 & 0.5136 & \textbf{0.5628} & \textbf{0.5062} & 0.0614 \\
Label seed 43 & 0.6284 & \textbf{0.6306} & 0.5972 & 0.6282 & \textbf{0.5832} & 0.0657 \\
\midrule
Mean & 0.5906 & 0.5917 & 0.5554 & \textbf{0.5955} & \textbf{0.5447} & 0.0636 \\
\bottomrule
\end{tabular}
\end{table}

\newpage

\section*{D. More details on the algorithms}


Algorithm ~\ref{alg:gaussian-ring} provides the pseudo-code of the Gaussian ring, and Algorithm~\ref{alg:scm-training-short} provides the pseudo-code summary of our proposed approach, namely Sparse Competitive Modules (SCM).

\begin{algorithm}[H]
\caption{Gaussian-ring soft module memberships}
\label{alg:gaussian-ring}
\begin{algorithmic}[1]
\Require Layer width $d$, number of modules $K$, temperatures $\tau_g,\tau_\sigma>0$ (usually 1.0),
budget factor $\beta$, width-floor fraction $\eta=0.1$, learnable logits $a\in\mathbb{R}^K$
\Ensure Membership matrix $G\in[0,1]^{d\times K}$ with $\sum_{k=1}^K G_{jk}=1$

\State Place fixed module centers on the unit ring:
\[
c_k \gets \frac{k-1}{K},\qquad k=1,\ldots,K.
\]
\State Set the total width budget:
\[
B_\sigma \gets \frac{\beta}{K}.
\]
\State Allocate widths under this budget:
\[
\sigma_{\min}\gets \eta\,\frac{B_\sigma}{K},
\qquad
\rho\gets \operatorname{softmax}(a/\tau_\sigma),
\]
\[
\sigma_k\gets \sigma_{\min}
+
(B_\sigma-K\sigma_{\min})\rho_k,
\qquad k=1,\ldots,K.
\]
\For{neuron $j=1,\ldots,d$}
    \State Set ring position:
    \[
    p_j\gets \frac{j-1}{d}.
    \]
    \For{module $k=1,\ldots,K$}
        \State Compute smooth periodic squared distance (for the ring effect):
        \[
        \Delta_{jk}^2
        \gets
        \frac{1-\cos(2\pi(p_j-c_k))}{2\pi^2}.
        \]
        \State Compute Gaussian log-affinity:
        \[
        \ell_{jk}
        \gets
        -\frac{\Delta_{jk}^2}{2\tau_g\sigma_k^2}.
        \]
    \EndFor
    \State Normalize affinities across modules:
    \[
    G_{j,:}\gets \operatorname{softmax}(\ell_{j,:}).
    \]
\EndFor
\State \Return $G$
\end{algorithmic}
\end{algorithm}

\begin{algorithm}[t]
\caption{Sparse Competitive Modules (SCM) training loop}
\label{alg:scm-training-short}
\begin{algorithmic}[1]
\Require Data $\mathcal{D}$, network $f_\theta$, routed layers $\mathcal{L}$,
routing temperature $\tau_r$, routing weight $\alpha$
\Ensure Trained parameters $\theta$ and Gaussian ring logits $\{a^\ell\}_{\ell\in\mathcal{L}}$

\For{each training batch $(x,y)$}
    \State Run the forward pass of $f_\theta$, applying the following update at each routed layer $\ell\in\mathcal{L}$:
    \Statex

    \For{each routed layer $\ell\in\mathcal{L}$}
        \State Compute pre-activation:
        $
        u^\ell\gets W^\ell h^{\ell-1}.
        $

        \State Compute module memberships:
        \[
        G^\ell\gets
        \textsc{GaussianRing}
        (d_\ell,K_\ell,\tau_g^\ell,\tau_\sigma^\ell,\beta^\ell,a^\ell).
        \]

        \State Compute module energies:
        \[
        E^\ell_{bk}
        \gets
        \left\|
        \operatorname{ReLU}(u^\ell_b)\odot G^\ell_{:,k}
        \right\|_2+\varepsilon.
        \]

        \State Compute routing probabilities:
        \[
        q^\ell_b\gets \operatorname{softmax}(E^\ell_b/\tau_r).
        \]

        \State Select the winning module:
        \[
        z^\ell_b\gets
        \operatorname{stopgrad}\!\left(\arg\max_k E^\ell_{bk}\right).
        \]

        \State Apply sparse top-1 routing:
        \[
        h^\ell_b
        \gets
        \operatorname{ReLU}(u^\ell_b)\odot G^\ell_{:,z^\ell_b}.
        \]

        \State Compute average module usage:
        \[
        \bar q^\ell_k
        \gets
        \frac{1}{|\mathcal{B}|}
        \sum_{b=1}^{|\mathcal{B}|}q^\ell_{bk}.
        \]

        \State Compute routed-layer loss:
        \[
        \mathcal{L}^\ell_{\mathrm{route}}
        \gets
        -
        \frac{1}{|\mathcal{B}|}
        \sum_{b=1}^{|\mathcal{B}|}
        \sum_{k=1}^{K_\ell}
        q^\ell_{bk}\log q^\ell_{bk}
        +
        \sum_{k=1}^{K_\ell}
        \bar q^\ell_k\log \bar q^\ell_k.
        \]
    \EndFor
    \State Compute total loss:
    \[
    \mathcal{L}
    \gets
    \operatorname{CE}(f_\theta(x),y)
    +
    \alpha
    \sum_{\ell\in\mathcal{L}}
    \mathcal{L}^\ell_{\mathrm{route}}.
    \]
    \State Update $\theta$ and $\{a^\ell\}_{\ell\in\mathcal{L}}$ by backpropagation.
\EndFor
\end{algorithmic}
\end{algorithm}